\documentclass[sigconf]{acmart}
\setcopyright{none}
\makeatletter
\renewcommand\@formatdoi[1]{} 
\makeatother
\renewcommand\footnotetextcopyrightpermission[1]{} 

\acmConference[]{}{}{}
\acmISBN{}
\acmDOI{}

\usepackage{algorithm}
\usepackage{multirow}
\usepackage{subcaption} 
\usepackage{booktabs} 
\usepackage{amsmath}  
\usepackage{tabularx}
\usepackage{adjustbox}
\usepackage{longtable}

\begin{document}

\title{ReFilter: Improving Robustness of Retrieval-Augmented Generation via Gated Filter}
\author{Yixin Chen}
\authornote{Both authors contributed equally to this research.}
\email{yixichen-c@my.cityu.edu.hk}
\affiliation{%
  \institution{City University of Hong Kong}
  \city{Hong Kong}
  \country{China}
}

\author{Ying Xiong*}
\email{ying.xiong@mbzuai.ac.ae}
\affiliation{%
  \institution{MBZUAI}
  \city{Abu Dhabi}
  \country{UAE}
}

\author{Shangyu Wu}
\email{shangyu.wu@mbzuai.ac.ae}
\authornote{Corresponding author.}
\affiliation{%
  \institution{MBZUAI}
  \city{Abu Dhabi}
  \country{UAE}
}

\author{Xiangrui Ke}
\email{x3ke@uwaterloo.ca}
\affiliation{%
 \institution{University of Waterloo}
 \city{Waterloo}
 \country{Canada}
 }

\author{Nan Guan}
\email{nanguan@cityu.edu.hk}
\affiliation{%
  \institution{City University of Hong Kong}
  \city{Hong Kong}
  \country{China}
}

\author{Chun Jason Xue}
\email{jason.xue@mbzuai.ac.ae}
\affiliation{%
  \institution{MBZUAI}
  \city{Abu Dhabi}
  \country{UAE}
}


\begin{abstract}
Retrieval-augmented generation (RAG) has become a dominant paradigm for grounding large language models (LLMs) with external evidence in knowledge-intensive question answering. 
A core design choice is how to fuse retrieved samples into the LLMs, where existing internal fusion approaches broadly fall into query-based fusion, parametric fusion, and latent-based fusion.
Despite their effectiveness at modest retrieval scales, these methods often fail to scale gracefully as the number of retrieved candidates $k$ increases: 
Larger $k$ improves evidence coverage, yet realistic top-$k$ retrieval inevitably contains irrelevant or redundant content and increases the inference cost.

To address these limitations, we propose \textbf{ReFilter}, a novel latent-based fusion framework that performs token-level filtering and fusion. 
ReFilter consists of three key components: a context encoder for encoding context features, a gated filter for weighting each token, and a token fusion module for integrating the weighted token feature into the LLM's hidden states. 
Our experiments across four general-domain QA benchmarks show that ReFilter consistently achieves the best average performance under both in-domain adaptation and out-of-domain transfer.
ReFilter further generalizes to five biomedical QA benchmarks in zero-shot transfer without domain fine-tuning, reaching 70.01\% average accuracy with Qwen2.5-14B-Instruct.


\end{abstract}

\maketitle
\section{Introduction}
\label{sec:intro}

Large language models (LLMs) serve as a powerful foundation for question answering (QA)~\cite{20nips_fewshot}, yet their knowledge is fundamentally constrained by their training data, internalization processes, and fixed cut-off dates~\cite{21nips_mg, 22acl_tm, 24emnlp_hallu}. 
This limitation often leads to fluent but ungrounded or incorrect answers, particularly for queries requiring long-tail, evolving, or specialized knowledge~\cite{24emnlp_hallu}. 
Retrieval-Augmented Generation (RAG) mitigates this by grounding LLM generation in dynamically retrieved external evidence~\cite{20icml_realm, 20nips_ragnlp, 24arxiv_nlprag}. 
However, how to effectively and efficiently integrate the retrieved knowledge into LLMs remains a central challenge.

Existing internal fusion approaches can be classified into three categories: 
\textit{Query-based Fusion} for concatenating samples with the input~\cite{20nips_ragnlp, 24naacl_replug}, 
\textit{Parametric Fusion} for encoding samples into LoRA-like parameters~\cite{prag,dyprag}, 
and \textit{Latent-based Fusion} for integrating features into hidden states of LLMs~\cite{24iclr-refusion, 22icml-retro}.
Although these fusion methods can all effectively inject external knowledge into LLMs, they do not scale well with more retrievals containing evidence. 
The major reason is that in realistic retrieval, the top-$k$ results often contain irrelevant, redundant, or even conflicting content, and prior work shows that LLMs can be distracted by such noise, leading to degraded generation quality~\cite{25acl_distracting, D24tacl_lim, 23icml_irrllm, 24iclr_rag_irr}. 
Besides, a larger $k$ will also lead to a higher inference cost.
How to maintain high inference efficiency while improving performance with a large number of retrievals becomes the key to RAG.


To address the above limitations, we propose \textbf{ReFilter}, a novel latent-based fusion framework that performs token-level filtering and fusion.
Our ReFilter is a plug-and-play module comprising three submodules: a context encoder, a gated filter, and a token fusion module.
First, the context encoder transforms the retrieved chunks into a flattened context feature.
Then, the gated filter computes the importance scores for each token in the context.
Finally, the token fusion module integrates weighted token features into the hidden states of the LLM's specific layer.
With those modules, our ReFilter can not only maintain high efficiency without feeding long-context input sequences and by enabling batching, but also improve robustness by reducing the impact of irrelevant or noisy tokens.



We evaluate our ReFilter under two realistic deployment regimes: In-Domain Adaptation, where the refilter is trained on limited task-specific data, and Out-of-Domain Transfer, where a refilter trained on general data is applied directly to specialized domains like clinical QA without further fine-tuning. 
This comprehensive evaluation demonstrates our method's effectiveness, efficiency, and strong generalization across domains.

The main contributions of this paper are:
\begin{itemize}
    \item We identify a key scalability bottleneck in RAG: as $k$ increases, noisy (irrelevant/redundant/conflicting) retrieval can decouple higher evidence recall from downstream QA performance.
    \item We propose \textbf{ReFilter}, a latent-based fusion module that performs token-level filtering and efficient hidden-state fusion to suppress noise without long-context prompting.
    \item Extensive experiments under in-domain adaptation and out-of-domain transfer show consistent improvements over strong baselines, including robust zero-shot generalization to biomedical QA.
\end{itemize}
\section{Background and Motivation}
\label{back}

\begin{figure}[t]
    \centering
    \includegraphics[width=\linewidth]{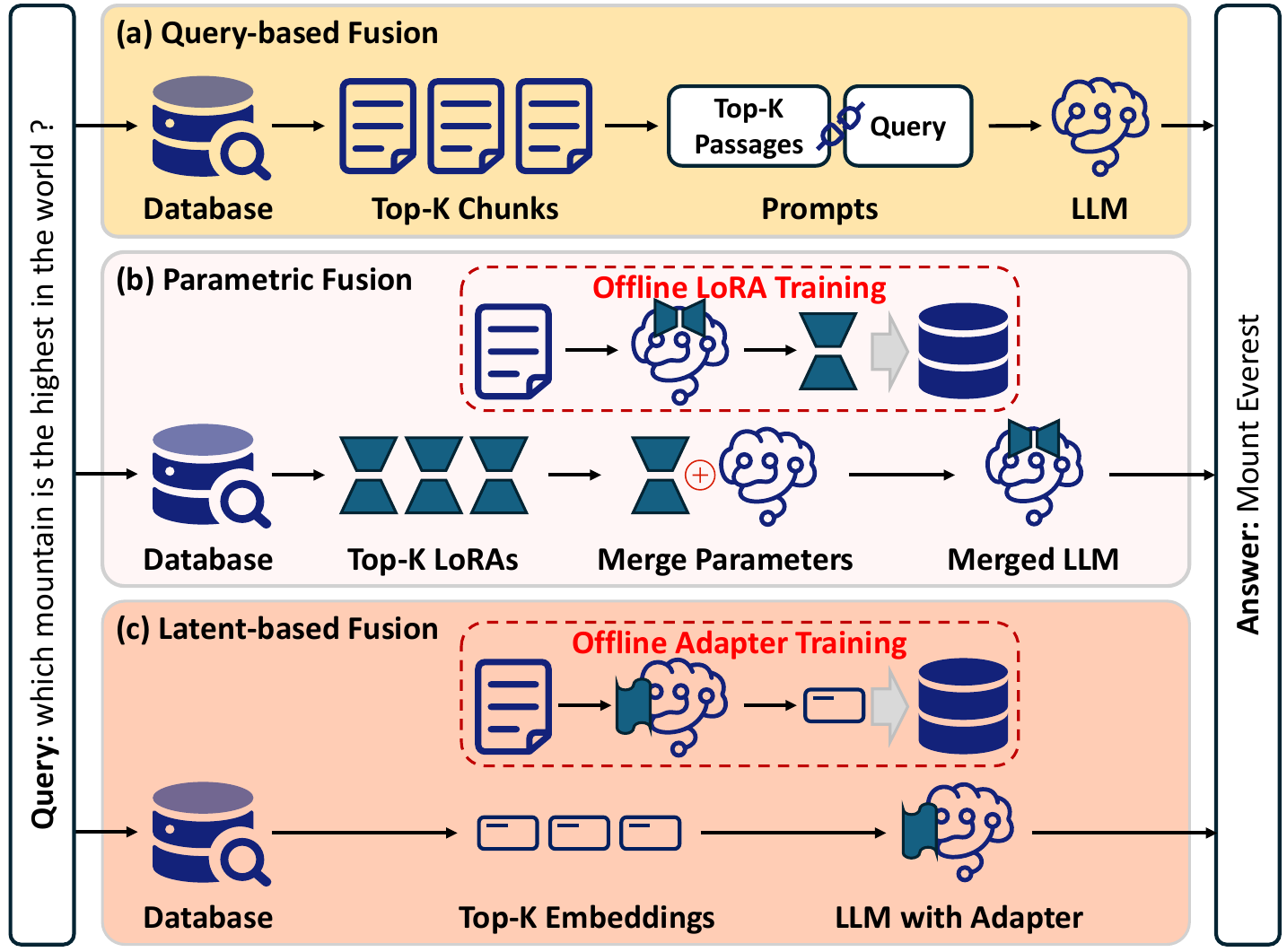}
    \caption{Three retrieval fusion paradigms in RAG: query-based fusion, parametric fusion, and latent-based fusion.}
    \label{fig:fusion_taxonomy}
\end{figure}

\subsection{Retrieval Fusions}
Retrieval-augmented generation (RAG) enhances large language models (LLMs) by integrating external knowledge into the generation process, which has become a dominant paradigm for knowledge-intensive tasks~\cite{20nips_ragnlp}.
Formally, given a query $q$, a retriever will search an external database for the top-$k$ related chunks as external evidence, and the LLM generates the answer $a$ conditioned on $q$ and the retrieved evidence.

The key to RAG is how to integrate evidence into the LLM generation process.
As shown in Figure~\ref{fig:fusion_taxonomy}, existing approaches can be categorized into \textbf{Query-based Fusion}, \textbf{Parametric Fusion}, and \textbf{Latent-based Fusion}.
Query-based fusion concatenates retrieved chunks into the input, leveraging LLMs' in-context learning capabilities~\cite{20nips_ragnlp, 24naacl_replug}.
Parametric fusion encodes external knowledge into parameters (e.g., LoRA modules) and then augments the LLMs by merging those parameters of top-$k$ related evidence~\cite{prag,dyprag}.
Different from parametric fusion, latent-based fusion encodes external knowledge into feature embeddings and then fuses them into LLMs via lightweight adapters~\cite{24iclr-refusion}.

\begin{figure}[t]
    \centering
    \begin{subfigure}[b]{0.23\textwidth}
        \centering
        \includegraphics[width=\linewidth]{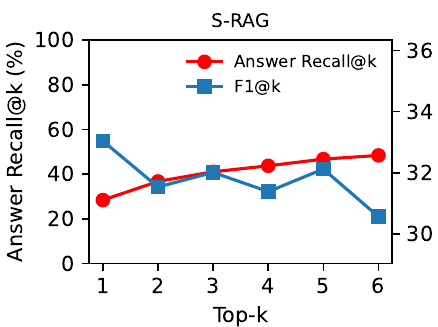}
        \caption{llama3.2-1B}
    \end{subfigure}
    \hfill
    \begin{subfigure}[b]{0.23\textwidth}
        \centering
        \includegraphics[width=\linewidth]{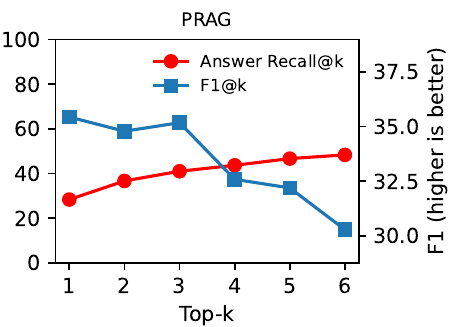}
        \caption{llama3.2-1B}
        
    \end{subfigure}

    \begin{subfigure}[b]{0.23\textwidth}
        \centering
        \includegraphics[width=\linewidth]{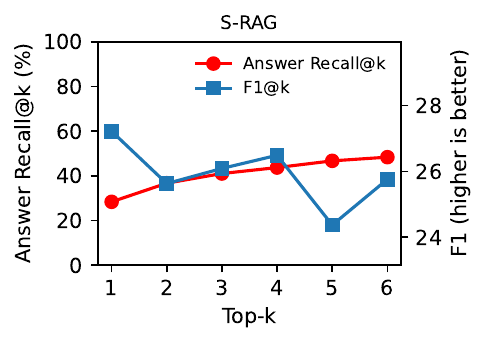}
        \caption{Qwen2.5-1.5B}
    \end{subfigure}
    \hfill
    \begin{subfigure}[b]{0.23\textwidth}
        \centering
        \includegraphics[width=\linewidth]{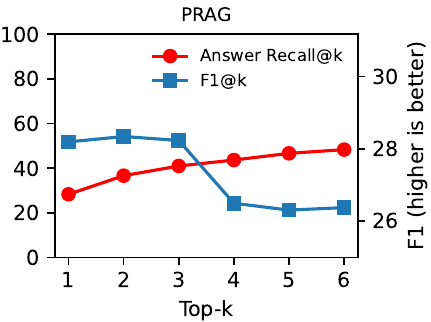}   
        \caption{Qwen2.5-1.5B}
    \end{subfigure}  
    
    \caption{The top-k recall and performance based on different backbones.}
    \label{fig:motivation}
\end{figure}

\subsection{Motivation}
To improve RAG performance, a simple approach is to increase $k$: more candidates increase the likelihood that the gold evidence is retrieved.
However, we observe a counterintuitive phenomenon when scaling the amount of retrieved context. 
In our preliminary study shown in Figure~\ref{fig:motivation}, we gradually increase the number of retrieved samples (and correspondingly the number of evidence-containing samples). 
As expected, the evidence recall increases monotonically with $k$, indicating improved coverage of gold evidence.
Yet we observe a clear \emph{decoupling} between retrieval recall and downstream performance: although recall keeps increasing, task performance (e.g., accuracy) improves only up to a turning point and then deteriorates as $k$ further grows.

The reason for this phenomenon is simple but consequential: \textit{larger $k$ brings in not only additional relevant evidence, but also more irrelevant, redundant, and sometimes conflicting chunks.} 
Query-based fusion requires LLMs to recognize and then ignore those irrelevant chunks.
Parametric fusion may directly pollute the original LLM parameters when injecting the parameters of those irrelevant chunks.
Latent-based fusion usually adopts a soft-fusion way to integrate features into LLMs' hidden states, which might be vulnerable to irrelevant features.
These limitations suggest that the degradation is not due to a specific implementation, but rather a systematic challenge of retrieval fusion under noisy evidence.

Finally, among the fusion families, we choose to optimize latent-based fusion for efficiency and modularity.
Compared with query-based fusion, latent-based fusion avoids long input sequences, thereby mitigating the quadratic attention cost with respect to input length.
Parametric fusion cannot process queries in batches, as it cannot merge the parameters of different queries simultaneously.
Besides, latent-based fusion offers great flexibility to design different lightweight adapters that filter those irrelevant features before injecting them into the LLMs.

\section{Methodology}
\label{medthod}

\begin{figure*}[t]
  \centering
  \includegraphics[width=0.8\linewidth]{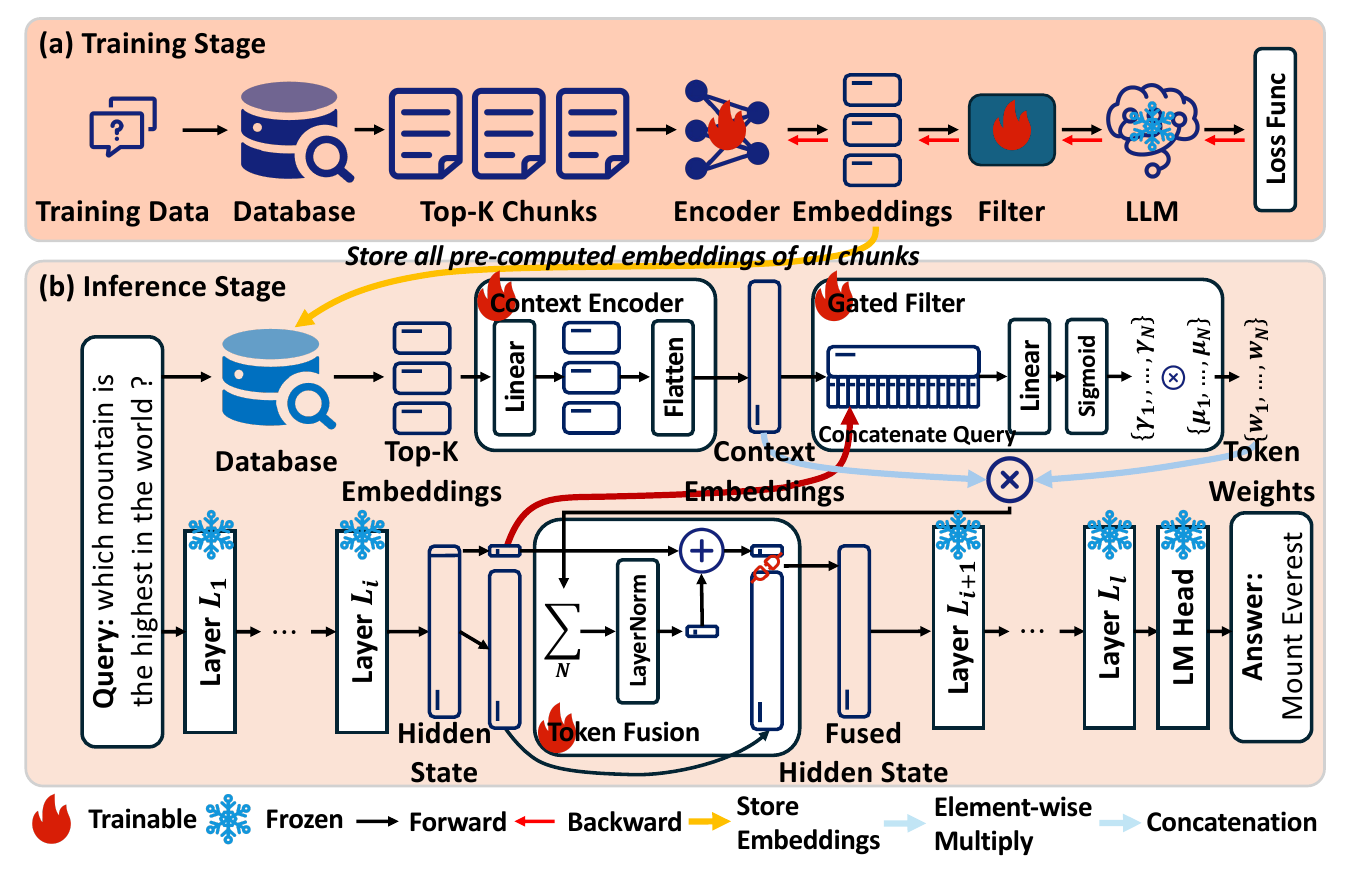}
  \caption{Overview of our ReFilter.}
  \label{fig:overview}
\end{figure*}

\subsection{Overview}
We study the retrieval-augmented question answering (QA) where an LLM answers a batch of queries $Q$ with the help of an external corpus $\mathcal{C}$.
A retriever returns top-$k$ chunks for each query $q_b$.:
\begin{equation}
\mathcal{C}_k(q_b)=\{c_{b,1},\dots, c_{b,i},\ldots,c_{b,k}\}
\end{equation}
Unlike prompt-based RAG that concatenates retrieved text into the input, we keep the prompt short and fuse retrievals at the representation level.
Specifically, we introduce a filter component that includes a context encoder to build the token pool ($k \times \texttt{chunk\_len}$ tokens), a gated filter for filtering irrelevant tokens, and a token fusion for injecting token representations into the LLM's hidden states.
The filter component will inject the retrieval representations into a specific layer of the LLM.

\paragraph{Notations.}
Let $\mathbf{H}\in\mathbb{R}^{B\times P\times L\times d_m}$ denote all hidden states of the backbone LLM, where $B, P, L, d_m$ are the batch size, input sequence length, number of layers and dimension size.
We use $p_b$ to denote the injected layer index for the $b$-th query during the whole decoding stage. 
We collect these positions in a tensor $\mathbf{p}\in\mathbb{N}^{B}$. For a complete list of notations used in this paper, see \S\ref{notations} in the Appendix.


\subsection{Trainable Context Encoder for Retrieved Chunks}
\label{sec:textencoder}
Given a batch of queries $Q=\{q_1, \ldots, q_B\}$, we first retrieve top-$k$ chunks for each query and aim to create the context embeddings that are semantically similar to the LLM's hidden state space.
To achieve this, we leverage an external Transformer encoder $E(\cdot)$ (e.g., a BERT-like model) to encode retrieved chunks and build a context encoder to project the features into the LLM's hidden space.
Concretely, for the $i$-th chunk $c_{b,i}$ of $b$-th query, we tokenize and pad/truncate to length $s$ and obtain text feature $f_{b,i}$:
\begin{equation}
f_{b,i} = E(c_{b,i})\in\mathbb{R}^{s\times d_e},
\end{equation}
where $d_e$ is the feature dimension size.
Then, the context encoder transforms the text feature $f_{b,i}$ into the context feature $C_{b,i}$ for the LLM via a learnable linear projection module with parameters $W_p\in\mathbb{R}^{d_e\times d_{m}}$:
\begin{equation}
C_{b,i} = W_p^{\top} f_{b,i} \in \mathbb{R}^{s\times d_{m}}.
\label{eq:proj}
\end{equation}
Then, we merge all $k$ chunks and then flatten each query's retrieval dimension to obtain the \textbf{context embeddings} $C \in \mathbb{R}^{B\times N\times d_{m}}$, where $N = k \times s$.
To improve the inference efficiency, we can cache the precomputed text feature $f_{b,i}$ when the corpus is static.
We use token-level features rather than sentence-level features, enabling fine-grained filtering. 

\subsection{Gated Filter for Token-level Weighting}
\label{sec:gatedlayer}
The gated filter module aims to produce the importance weights for each token.
It takes (1) the output hidden states $\mathbf{H}$ of the injected layer of the LLM and (2) the context embeddings $C$ as inputs, and returns an array of token weights $W_t\in\mathbb{R}^{B\times N}$. 

\subsubsection{Decision State}
For the $b$-th each query, we extract the LLM's output hidden states $h_b$ of the last token at the injection layer $p_b$:
\begin{equation}
h_b = \mathbf{H}_{b,p_b,L, :} \in \mathbb{R}^{d_m}.
\label{eq:q}
\end{equation}
This vector serves as a \textbf{query-conditioned decision state} that guides evidence filtering.

\subsubsection{Token-wise Dynamic Gate}
We compute a per-token gate for each token in context embeddings $C_{b,j}\in\mathbb{R}^{d_m}, j=1,\ldots, N$.
We first concatenate the token embedding with the decision state:
\begin{equation}
g_{b,j} = [C_{b,j}; h_b] \in \mathbb{R}^{2d_m}.
\end{equation}
Then, a lightweight linear scorer with a sigmoid activation function is used to produce the \textbf{dynamic gate}:
\begin{equation}
\gamma_{b,j} = \sigma\left(W_g^\top g_{b,j} + a_g\right),
\quad 0 < \gamma_{b,j} < 1,
\label{eq:dynamic_gate}
\end{equation}
where $W_g\in\mathbb{R}^{2d_m}$ and $a_g\in\mathbb{R}$ are learnable parameters.

\subsubsection{Learnable Position Mask}
Although the dynamic gate $\gamma$ captures the query-conditioned semantic importance of retrieved tokens, it does not explicitly model the positional effects in the evidence sequence.
Therefore, we introduce a learnable \textbf{position mask} $\mu=\{\mu_1, \ldots, \mu_N\}, \mu_j\in\mathbb{R}$ to encode position-dependent importance.
These parameters are shared across all queries and initialized to ones.

Because the flattening operation is performed on the sequence dimension (across different retrievals),
$\mu_j$ can capture global priors, e.g., retrieved chunks with high retrieval ranking are generally more useful; chunks in the title position might be more important, etc.
Unlike a purely dynamic gate, this position mask can encode dataset-level structural biases.

\subsubsection{Token Weights via Aggregating Dynamic Gate and Position Mask}
To get the token weights, we combine the dynamic gate and position mask by performing element-wise multiplication:
\begin{equation}
W_{t} = \{\mu_j \cdot \gamma_{b,j}\}, W_t\in\mathbb{R}^{B\times N}.
\label{eq:final_weight}
\end{equation}
These token weights are computed on the fly for every token of each query.
Unlike listwise softmax attention, our weights are \emph{not} constrained to sum to one.
This design allows multiple evidence tokens to contribute simultaneously (independent gating).
Finally, we compute the weighted token features by performing element-wise multiplication over context embeddings and token weights:
\begin{equation}
    \hat{C}=W_t\cdot C \in\mathbb{R}^{B\times N\times d_m}
\end{equation}

\subsection{Token Fusion}
\label{sec:injection}
After obtaining the weighted token features $\hat{C}$, we update the LLM's hidden states by adding the scaled context feature $\alpha *r_b$ onto the injected layer's spec hidden state: $\mathbf{H}_{b,p_b,L,:}$: 
\begin{equation}
    r_b=\text{LayerNorm}\left(\sum^N_{j=1} \text{Dropout}(\hat{C}_{b,j}) \right)
\end{equation}
\begin{equation}
    \hat{\mathbf{H}}_{b,p_b,L,:}=\mathbf{H}_{b,p_b,L,:} + \alpha \cdot r_b
\end{equation}
where $\alpha$ is a learnable scalar that controls the overall strength.
If retrieval is unavailable (or we don't inject), the layer bypasses and returns the original hidden states with zero auxiliary loss.




\subsection{Training Objective}
\label{sec:loss}
We train the fusion module with supervised QA data using standard teacher forcing.
Let $y=[y_1,\dots,y_T]$ be the target answer tokens, and let the backbone LM head map the updated last-layer states to token logits.
We minimize the negative log-likelihood:
\begin{equation}
\mathcal{L}_{\mathrm{NLL}} = -\sum_{t=1}^{T}\log p(y_t \mid y_{<t}, q, \mathcal{C}_k(q)).
\end{equation}

\paragraph{Gate sparsity regularization.}
To encourage selective evidence usage, we add a sparsity penalty on the dynamic gates:
\begin{equation}
\mathcal{L}_{\mathrm{gate}} = \frac{1}{BN}\sum_{b=1}^{B}\sum_{j=1}^{N}\gamma_{b,j},
\label{eq:gate_sparsity}
\end{equation}
The final objective is:
\begin{equation}
\mathcal{L} = \mathcal{L}_{\mathrm{NLL}} + \lambda \mathcal{L}_{\mathrm{gate}},
\label{eq:final_loss}
\end{equation}
where $\lambda$ balances answer accuracy and evidence selectivity.



\section{Experiment Settings}
\label{exp}


\subsection{Deployment Regimes}
\label{sec:protocol}
We evaluate under two deployment regimes:
\noindent\textbf{In-domain adaptation.}
We train the retrieval filter on the \emph{training} split of the target dataset and report performance on the held-out \emph{test} split.
This protocol reflects a practical setting where limited in-domain supervision is available for adapting the retrieval utilization module
without modifying the backbone LLM.
\noindent\textbf{Out-of-domain transfer.}
We train the retrieval filter on a large-scale \emph{general} dataset consisting of 150,000 diverse instruction-output pairs, drawn from Alpaca~\cite{alpaca}, KILT~\cite{kilt}, ASQA~\cite{asqa}, and OpenBookQA~\cite{openbook}.
Then we evaluate \emph{without any downstream training} on:
(i) general QA benchmarks and (ii) biomedical QA benchmarks.
This measures cross-domain generalization and robustness to domain shift.

It is noted that for both deployment regimes, we use Wikipedia as the external retrieval database.

\subsection{Datasets}
\label{sec:datasets}

We evaluated our method using QA datasets from two domains: general and biomedical. The general domain encompasses multi-hop reasoning and commonsense reasoning. Specifically, following PRAG~\cite{prag}, we selected the following datasets: 2WikiMultihopQA (2WQA), HotpotQA (HPQA), PopQA, and ComplexWebQuestions (CWQ). Our reported results are evaluated the total sub-task data.

For the biomedical domain, following Xiong et al.~\cite{24acl_bench_medical_rag}, we performed evaluations on MedQA~\cite{medqa}, MedMCQA~\cite{medmcqa}, PubMedQA~\cite{pubmedqa}, BioASQ~\cite{bioasq1}, and MMLU\_Med~\cite{mmlu}. MMLU\_Med, MedQA-US, and MedMCQA are medical examination QA datasets, while PubMedQA and BioASQ-Y/N are biomedical research QA datasets. The MMLU\_Med dataset consists of six biomedical-related tasks, including anatomy, clinical knowledge, professional medicine, human genetics, college medicine, and college biology.

\subsection{Backbone LLM}
\label{sec:backbone}
In our experimental setup, we employed multiple backbone models from different families and scales to ensure a comprehensive evaluation. Specifically, we selected the following models: LLaMA-3.2-1B-Instruct, LLaMA-3-8B-Instruct, Qwen2.5-1.5B-Instruct, and Qwen2.5-14B-Instruct. This selection spans a range of parameter sizes—from 1 billion to 14 billion parameters—and covers two prominent model families (LLaMA and Qwen), allowing us to examine both model-scale effects and architectural variations across distinct instruction-tuned lineages.

\subsection{Retrieval Corpus and Indexing}
\label{sec:corpus}
We build a retrieval corpus by chunking documents into chunks of fixed length (up to \texttt{chunk\_len} tokens).
For each query, we retrieve top-$k$ passages and form an retrieval token pool of size $N=k\times\texttt{chunk\_len}$.
For both general QA and biomedical QA, we used wikipedia as retrieval corpus and we used BM25~\cite{bm25} as indexing to retrieve top-$k$ candidates.

\subsection{Baselines}
\label{sec:baselines}
We compare against the following baselines:
(1) Vanilla LLM. The backbone LLM answers questions without external retrieval. (2) In-context standard RAG (S-RAG). We retrieve top-$k$ passages and append them to the prompt (retrieve-then-read) \cite{20nips_ragnlp}.
This baseline tests the conventional long-context knowledge injection.
(3) PRAG~\cite{prag}. It reduces dependency on input context length by injecting relevant documents into the parameters of the LLM through an offline LoRA adapter.
(4) DyPRAG~\cite{dyprag}. It employs the same offline one-to-one encoding process as PRAG and further incorporates a hypernetwork that transforms documents into the format required by LoRA adapters during online inference.

\subsection{Training Details}
\label{sec:training}
We train the filter retrieval parameters with teacher-forced next-token prediction on QA pairs using cross-entropy loss:
$\mathcal{L}_{\text{NLL}} = -\sum_t \log p(y_t \mid y_{<t}, q, \mathcal{E})$,
where $\mathcal{E}$ denotes the retrieved token pool.
We optionally regularize the token-importance distribution to encourage sparsity (e.g., entropy penalty).
Training is conducted for $8$ epochs using the AdamW optimizer with a learning rate of $1e^{-5}$, a batch size of 16 and a $5\%$ warmup ratio. 
We freeze the original LLM parameters and only fine-tune the text encoder and the lightweight gated filter inserted into the last 3 transformer layers.
Hyperparameter tuning on the development set selects $k=3$ retrieved chunks with chunk length of 256 tokens from $\{1,3,5\}$ and $\{128,256,512\}$ respectively.
We also tune the gating threshold (or gating temperature) to balance retrieval usage and robustness.

\subsection{Evaluation Metrics}
\label{sec:metrics}
We report standard QA metrics for each dataset. For the general QA dataseet, we report F1 scores as the evaluation metric.
And for biomedical QA, we report accuracy.
We also report efficiency metrics including average input length (tokens), time-to-first-token (TTFT),
and stable generation token per second/throughput (tokens/second).

\section{Results}
\label{res}

\begin{table*}[t]
\centering
\caption{In-domain and out-of-domain results. In-domain: train on each dataset's training split and test on its test split. Out-of-domain: train on a general instruction dataset and test on each downstream dataset's test split.}
\label{tab:indomain-main}
\begin{tabular}{lcccccccccccc}
\toprule
\multirow{2}{*}{\textbf{Backbone}} & \multirow{2}{*}{\textbf{Method}} & \multicolumn{5}{c}{\textbf{In-Domain}} & \multicolumn{5}{c}{\textbf{Out-Of-Domain}} & \multirow{2}{*}{\textbf{AVG-all}} \\
\cmidrule(lr){3-7} \cmidrule(lr){8-12} & & \textsc{2WQA} & \textsc{HPQA} & \textsc{PopQA} & \textsc{CWQ} & \textbf{AVG} & \textsc{2WQA} & \textsc{HPQA} & \textsc{PopQA} & \textsc{CWQ} & \textbf{AVG}&\\
\midrule
\multirow{5}{*}{Llama-3.2-1B-Instruct} & Vanilla & 23.69&	19.05&	5.94&	34.78&	20.87&	-&	-&	-&	-&	-&	20.87  \\
& S-RAG & 22.38&	29.13&	20.29&	32.01&	25.95&	-&	-&	-&	-&	-&	25.95  \\
& PRAG & 24.47&	20.39&	23.07&	35.17&	25.78&	21.58&	18.79&	19.59&	31.21&	22.79&	24.28  \\
& DyPRAG &  23.67&	22.98&	12.56&	35.51&	23.68&	22.87&	20.98&	10.47&	32.32&	21.66&	22.67\\
& Ours & 26.87&	25.28&	23.38&	35.98&	27.88&	26.18&	22.46&	20.57&	33.48&	25.67&	\textbf{26.78}\\
\midrule
\multirow{5}{*}{Llama-3-8B-Instruct} & Vanilla &  31.99&	22.32&	14.27&	43.31&	27.97&	-&	-&	-&	-&	-&	27.97 \\
& S-RAG &  37.74&	24.19&	16.23&	43.11&	30.32&	-&	-&	-&	-&	-&	30.32  \\
& PRAG &  37.04&	33.12&	27.21&	42.06&	34.86&	32.19&	28.18&	24.10&	37.28&	30.44&	32.65  \\
& DyPRAG &  36.32&	28.26&	12.03&	42.33&	29.74&	32.85&	23.44&	10.15&	38.23&	26.17&	27.95  \\
& Ours& 38.42&	35.32&	27.38&	46.49&	36.90&	34.28&	32.19&	23.87&	42.75&	33.27&	\textbf{35.09} \\
\midrule
\multirow{5}{*}{Qwen2.5-1.5B-Instruct} & Vanilla &  22.09&	15.76&	7.06&	28.16&	18.27&	-&	-&	-&	-&	-&	18.27  \\
& S-RAG &  22.65&	16.22&	9.77&	26.09&	18.68&	-&	-&	-&	-&	-&	18.68  \\
& PRAG &  27.32&	16.16&	19.95&	28.23&	22.92&	23.18&	15.76&	13.64&	26.45&	19.76&	21.34  \\
& DyPRAG &  24.26&	19.37&	7.02&	30.28&	20.23&	21.86&	14.72&	5.21&	24.17&	16.49&	18.36  \\
& Ours& 29.47&	20.37&	18.38&	32.66&	25.22&	25.75&	17.54&	15.65&	26.57&	21.38&	\textbf{23.30} \\
\midrule
\multirow{5}{*}{Qwen2.5-14B-Instruct} & Vanilla &  32.16&	25.94&	24.84&	50.13&	33.27&	-&	-&	-&	-&	-&	33.27  \\
& S-RAG &  35.71&	26.28&	28.41&	46.64&	34.26&	-&	-&	-&	-&	-&	34.26  \\
& PRAG &  35.02&	27.48&	28.38&	47.29&	34.54&	30.12&	22.87&	26.19&	41.48&	30.17&	32.35  \\
& DyPRAG &  33.20&	26.59&	26.37&	45.87&	33.01&	29.19&	24.38&	23.38&	42.32&	29.82&	31.41  \\
& Ours& 39.23&	28.46&	30.37&	50.32&	37.10&	35.28&	28.36&	27.22&	45.87&	34.18&	\textbf{35.64} \\
\bottomrule
\end{tabular}
\end{table*}

\subsection{Main Results}
\label{sec:results-indomain}
We present a comprehensive evaluation of our proposed method against existing baselines across four diverse QA datasets (2WQA, HPQA, PopQA, CWQ) under both in-domain and out-of-domain settings. The results are summarized in Table~\ref{tab:indomain-main}. Overall, our method consistently achieves the highest average performance across all backbone models and evaluation scenarios, demonstrating strong generalization and robustness.
First, our approach outperforms all baselines in 10 out of 12 model-dataset combinations for in-domain evaluation, and in all 12 out-of-domain cases. On average, it achieves +2.54\% over the best baseline (PRAG) when averaged across models. Notably, with Llama-3-8B-Instruct, our method reaches an average score of 35.09, surpassing PRAG by 2.44\%.

Moreover, while PRAG shows reasonable cross-domain adaptation, our method exhibits significantly stronger out-of-domain performance. For example, with Qwen2.5-14B-Instruct, our method attains 35.28\% on 2WQA (out-of-domain), compared to PRAG’s 30.12\%—an improvement of 5.16\%. This highlights our method’s enhanced ability to generalize to unseen domains without fine-tuning.

Furthermore, our method scales effectively with model capacity. With the smallest model (Llama-3.2-1B-Instruct), we achieve 26.78\% average score, outperforming PRAG (24.28\%) and DyPRAG (22.67\%). With the largest (Qwen2.5-14B-Instruct), we reach 35.64\%, significantly ahead of PRAG (32.35\%) and Vanilla (33.27\%). This indicates that our approach is effective across both lightweight and large-scale models.

\subsection{Generalization on Biomedical Tasks}
\label{sec:results-ood}
We evaluate the zero-shot transfer capability of our method on five challenging medical QA benchmarks, with results shown in Table~\ref{tab:ood-main}. Models are trained on a general dataset and tested without any medical domain fine-tuning. Our method achieves the highest average score for every backbone model, demonstrating superior cross-domain knowledge transfer and robustness in specialized, knowledge-intensive domains.

Across all four model backbones, our method attains the best average score, outperforming the strongest baseline by up to +1.78\% (LLaMA-3.2-1B-Instruct) and establishing a new SOTA average of 70.01\% with the largest model, Qwen2.5-14B-Instruct. This consistent lead underscores the effectiveness of our approach in leveraging general training for specialized tasks.
Besides, our method delivers stable performance gains across diverse medical evaluation types—from USMLE-style exams (MEDQA) to biomedical research QA (PUBMEDQA, BIOASQ) and broad medical knowledge (MMLU\_MED)—confirming its robustness and general applicability within the medical domain.

\begin{table*}[t]
\centering
\caption{Medical QA transfer results: models are trained on a general QA dataset and evaluated zero-shot on medical QA benchmarks.}
\label{tab:ood-main}
\begin{tabular}{lccccccc}
\toprule
Backbone & Method & \textsc{medqa} & \textsc{medmcqa} & \textsc{pubmedqa} & \textsc{bioasq} & \textsc{mmlu\_med} & AVG. \\
\midrule
\multirow{5}{*}{LLaMA-3.2-1B-Instruct} & Vanilla &  25.14&	25.51&	30.20&	36.57&	31.96&	29.88  \\
& S-RAG &  25.22&	26.49&	35.20&	34.47&	32.97&	30.87 \\
& PRAG &  24.35&	28.38&	32.00&	35.44&	33.98&	30.83  \\
& DyPRAG &  25.14&	27.01&	28.40&	35.11&	33.24&	29.78  \\
& Ours  & 26.00&	28.50&	35.60&	37.22&	34.16&	\textbf{32.29}\\
\midrule
\multirow{5}{*}{LLaMA-3-8B-Instruct} & Vanilla &  45.72&	52.28&	51.80&	76.54&	69.88&	59.24 \\
& S-RAG &  49.96&	52.00&	54.20&	77.18&	67.22&	60.11  \\
& PRAG &  46.43&	52.43&	54.00&	76.05&	67.03&	59.19  \\
& DyPRAG &  44.70&	52.19&	54.80&	75.89&	67.95&	59.11  \\
& Ours&   51.06&	52.62&	56.00&	78.80&	69.79&	\textbf{61.65} \\
\midrule
\multirow{5}{*}{Qwen2.5-1.5B-Instruct} & Vanilla &  39.36&	36.00&	49.80&	66.18&	47.75&	47.82  \\
& S-RAG &  36.06&	37.29&	51.60&	66.18&	50.51&	48.33  \\
& PRAG &  38.49&	37.03&	51.80&	69.90&	52.25&	49.89 \\
& DyPRAG &  37.23&	36.34&	50.00&	67.96&	51.33&	48.57 \\
& Ours& 39.12&	37.75&	55.20&	71.04&	52.53&	\textbf{51.13} \\
\midrule
\multirow{5}{*}{Qwen2.5-14B-Instruct} & Vanilla &  66.22&	60.46&	54.00&	78.16&	81.27&	68.02  \\
& S-RAG &  67.64&	61.49&	55.20&	79.29&	81.63&	69.05  \\
& PRAG &  67.01&	61.15&	54.00&	79.13&	82.37&	68.73 \\
& DyPRAG &  67.87&	61.25&	55.20&	79.29&	81.73&	69.07  \\
& Ours& 67.79&	61.77&	56.80&	80.74&	82.92&	\textbf{70.01} \\
\bottomrule
\end{tabular}
\end{table*}

\subsection{Ablation Study}
\label{sec:results-ablation}
We conduct an ablation study using the Qwen2.5-1.5B-Instruct backbone to assess our core components (Table~\ref{tab:ablation}). The results demonstrate that each component contributes distinctly to performance, as the full model achieves the highest average score (25.22\%). Removing the token-level importance mask leads to the most substantial performance degradation ($-3.03\%$ on average), with a particularly sharp decline on 2WQA ($-7.54\%$). This result strongly confirms that fine-grained token filtering is essential to suppress noise in retrieved documents and prevent the model from being distracted by irrelevant content.
Removing the gating mechanism results in a modest average decline ($-0.81\%$), with notable drops on 2WQA and PopQA. This demonstrates the gate's role in dynamically balancing between the model's internal knowledge and external evidence, especially for fact-heavy queries.
Replacing token-level filter with chunk-level filter causes a consistent degradation ($-1.11\%$). While still beneficial compared to no masking, this shows that finer-grained, token-level evidence selection provides a clear advantage in forming precise contextual summaries.
In summary, the token-level filter is the primary driver for robustness, while the gating mechanism provides necessary adaptive control. Their combination enables both efficient and reliable knowledge integration.

\begin{table}[t]
\centering
\caption{Ablation study on representative benchmarks.}
\label{tab:ablation}
\begin{tabular}{lccccc}
\toprule
Variant & \textsc{2WQA} & \textsc{HPQA} & \textsc{PopQA} & \textsc{CWQ} & AVG. \\
\midrule
ReFilter &29.47&	20.37&	18.38&	32.66&	25.22  \\
w/o gating & 25.27&	23.67&	15.53&	33.17&	24.41 \\
w/o token mask & 21.93&	20.89&	15.48&	30.47&	22.19  \\
r.w. chunk filter & 24.89&	22.98&	17.39&	31.19&	24.11  \\
\bottomrule
\end{tabular}
\end{table}

\section{Analysis}
\label{sec:ana}

\subsection{Robustness to Noisy and Redundant Evidence}
\label{sec:results-robust}
\begin{figure}[t]
\centering
    \begin{subfigure}[b]{0.45\textwidth}
        \centering
        \includegraphics[width=\linewidth]{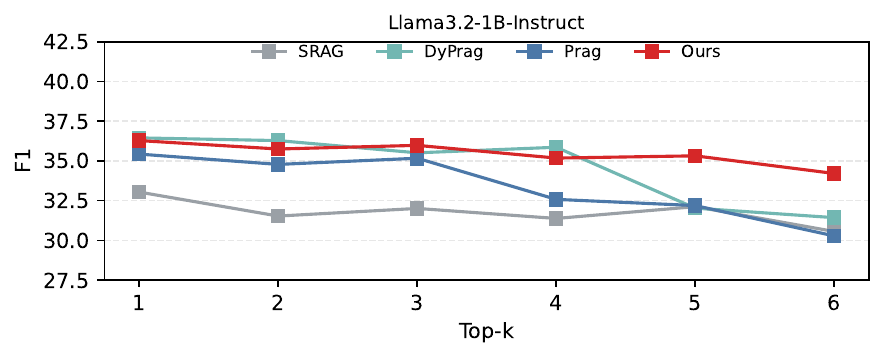}
    \end{subfigure}
    \hfill
    \begin{subfigure}[b]{0.45\textwidth}
        \centering
        \includegraphics[width=\linewidth]{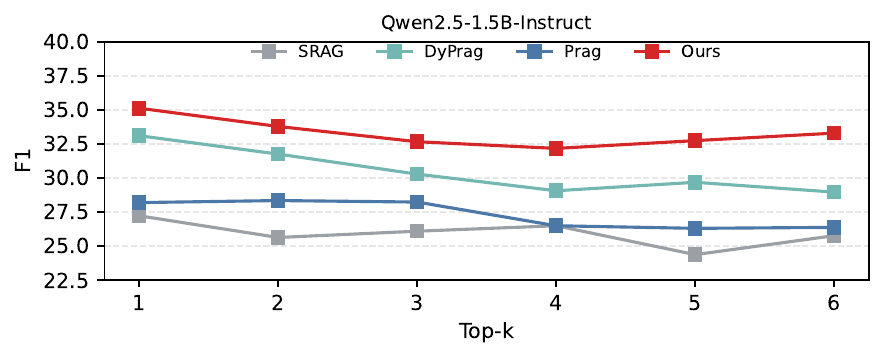}   
    \end{subfigure} 
\caption{Robustness to top-k.}
\label{fig:topk_noise}
\end{figure}

\begin{figure}[t]
\centering

    \begin{subfigure}[b]{0.23\textwidth}
        \centering
        \includegraphics[width=\linewidth]{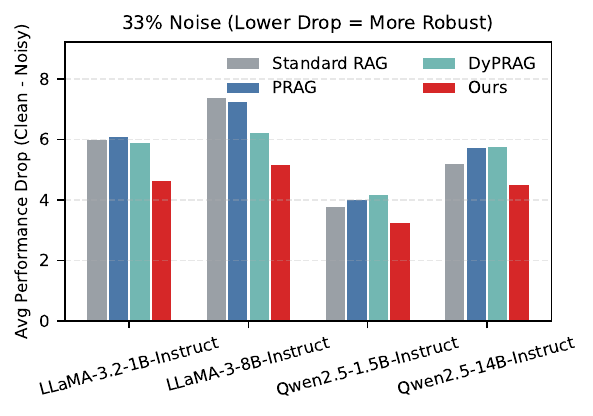}
    \end{subfigure}
    \hfill
    \begin{subfigure}[b]{0.23\textwidth}
        \centering
        \includegraphics[width=\linewidth]{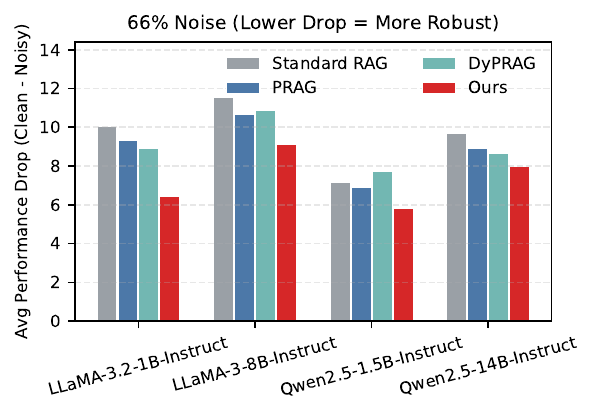}   
    \end{subfigure} 
\caption{Robustness to retrieval noise.}
\label{fig:noise}
\end{figure}

To systematically evaluate the robustness of our ReFilter against irrelevant or noisy retrieved content, we conduct two complementary analyses.
\paragraph{Varying the number of retrieved documents (Top-k)}
We first examine the sensitivity of models to different retrieval depths, varying the top-$k$ chunks retrieved by the retriever. We evaluated two backbones on the CWQ dataset. As shown in Figure~\ref{fig:topk_noise}, while baseline models such as Standard RAG, PRAG, and DyPRAG exhibit noticeable performance degradation or fluctuation as $k$ increases, our method consistently outperforms them across different top-$k$ values. This indicates that our fine-grained token-level reranking and gated integration mechanism effectively filters out marginally relevant or distracting tokens, making the model less sensitive to variations in retrieval quality.
\paragraph{Robustness under injected noise}
We further evaluate model robustness on the general QA by explicitly injecting irrelevant chunks into the retrieved set to simulate real-world retrieval errors. Specifically, we construct noisy input sets in which 33\% and 66\% of the retrieved content are randomly replaced with irrelevant text from medical Wikipedia. \footnote{https://huggingface.co/datasets/MedRAG/wikipedia}. 
As shown in Figure~\ref{fig:noise}, our model consistently exhibits the smallest performance drop across different backbone models compared to other methods. The robustness gains can be attributed to the gated fusion mechanism, which suppresses the influence of noisy tokens, and the learned token-level importance scores, which reduce reliance on fixed positional assumptions.

These two studies together highlight that our method maintains superior performance not only under ideal retrieval conditions but also when exposed to retrieval noise, showcasing its potential for real-world deployment where retrieval results may be imperfect.

\begin{figure}[t]
\centering
    \includegraphics[width=\linewidth]{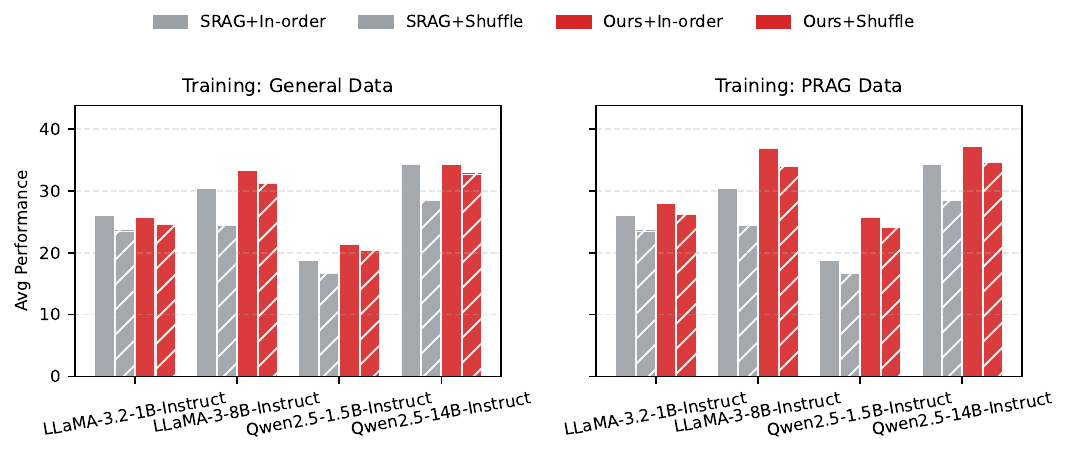}
    
\caption{Robustness to retrieval order.}
\label{fig:shuffle}
\end{figure}

\subsection{Robustness to Retrieval Order Perturbation}
To further assess the model’s robustness, we investigate its sensitivity to the ordering of retrieval. Specifically, we compare model performance when retrieved chunks are input in their original order (in-order) versus randomly shuffled. As shown in Figure~\ref{fig:shuffle}, our model exhibits minimal performance drop under shuffled inputs across all backbones, while standard RAG suffers more significantly. This suggests that our token-level reweighting and gated fusion mechanism effectively mitigates positional bias, allowing the model to focus on content relevance rather than surface order.

\begin{table}[t]
\centering
\caption{Inference latency (ms) across different batch sizes. PRAG and DyPRAG are only available at a batch size of 1.}
\label{tab:efficiency}
\begin{tabular}{lccccccc}
\toprule
Method & 1 & 4 & 8 & 16 & 32 & 64 \\
\midrule
PRAG & 339.94 & - & - & - & - & - \\
DyPRAG & 363.26 & - & - & - & - & - \\
S-RAG & 212.08 & 191.12 & 167.75 & 164.97 & 160.39 & 159.41 \\
Ours  & 279.72 & 192.13 & 155.95 & 148.28 & 131.73 & 128.81 \\
\bottomrule
\end{tabular}
\end{table}

\subsection{Efficiency: Context Cost and Latency}
\label{sec:results-efficiency}


We evaluate the latency and external storage costs of different fusion methods of Qwen2.5-1.5B-Instruct on the CWQ dataset. 
The latency results in Table~\ref{tab:efficiency} show that compared to parameteric fusion methods (e.g., PRAG and DyPRAG), standard RAG and our ReFilter are much faster when the batch size increases.
This indicates that both the standard RAG and our ReFilter can effectively utilize hardware parallelism, while parametric fusion methods usually require merging unique LoRAs for different queries.
When the batch size is small, our ReFilter achieves comparable efficiency to the standard RAG. 
This is because the extra time cost of inference on the Filter module offsets the efficiency gains.

For storage costs, PRAG requires about 6.24 GB for storing the LoRA weights. 
DyPRAG leverages a shared projection network to obtain LoRA weights for each query, thereby reducing storage costs to about 247 MB.
Our ReFilter stores precomputed token embeddings for each chunk, requiring an additional 675 MB of storage.
The results above show that our ReFilter can achieve a better trade-off between efficiency (latency/storage costs) and accuracy.

\subsection{Qualitative Analysis}
\label{sec:results-qual}
\begin{figure}[t]
\centering
    \begin{subfigure}[b]{0.46\textwidth}
        \centering
        \includegraphics[width=\linewidth]
        {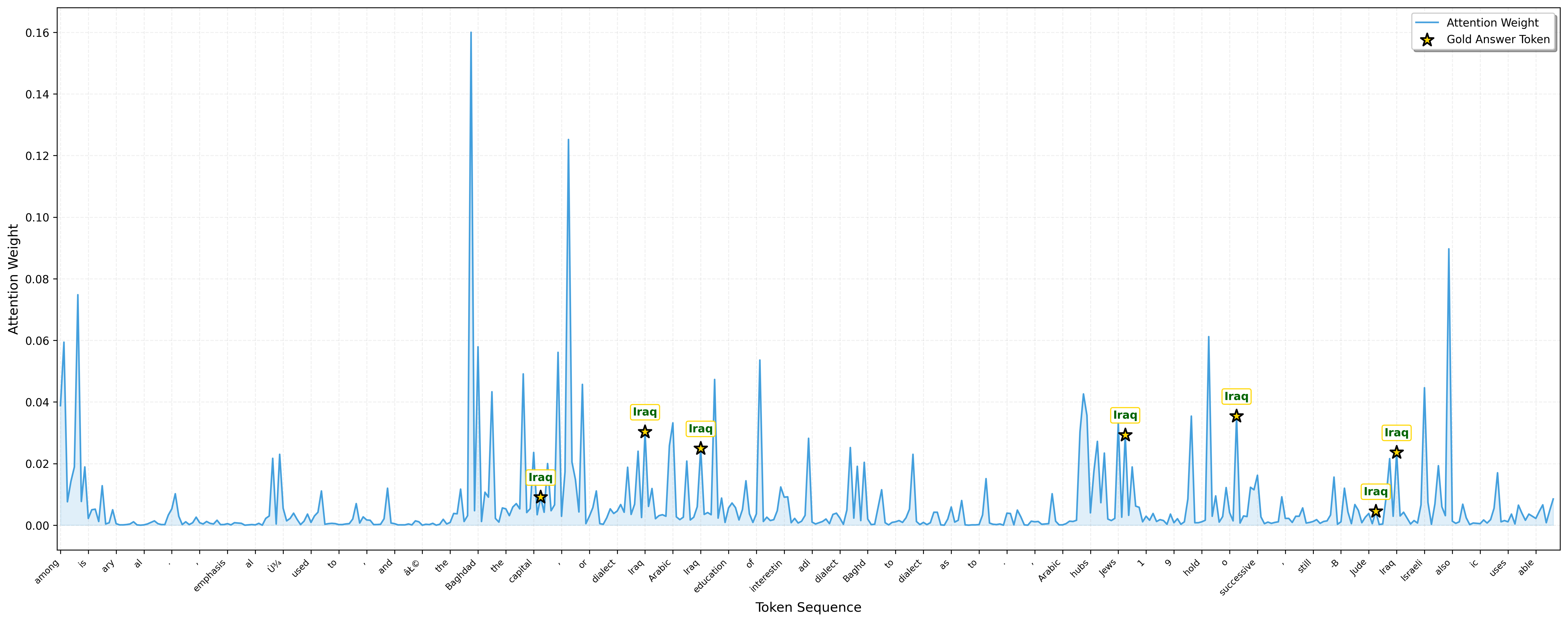}
    \end{subfigure}
    \hfill
    \begin{subfigure}[b]{0.46\textwidth}
        \centering
        \includegraphics[width=\linewidth]{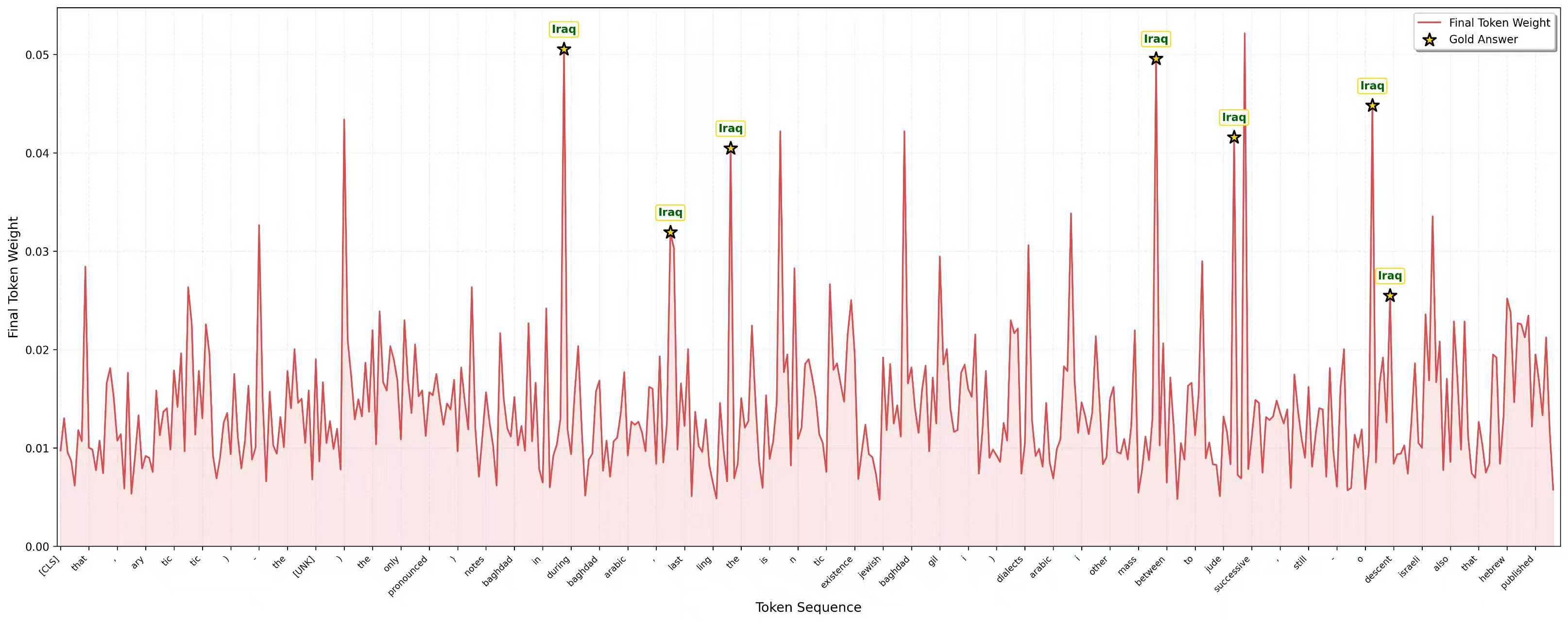}   
    \end{subfigure} 
\caption{Token-level final weights of our model (down) and Token-level attention weights of SRAG (upper). Stars denote occurrences of the gold answer token (“Iraq”).}
\label{fig:vis}
\end{figure}

We compare token-level visualizations of our model and the SRAG baseline on a QA example where the correct answer is “Iraq”. Figure~\ref{fig:vis} displays our model's final token weights and SRAG's attention weights. Gold-answer tokens are marked with stars.
Our model consistently assigns high weights to the answer token (“Iraq”), showing a clear focus on the spans that directly support the answer.
In contrast, SRAG's distribution is sharp, with a few tokens receiving extreme attention—often not aligned with the answer—while most tokens are near zero. Our model produces smoother, more stable weights that reliably highlight the correct evidence.
This case illustrates that our model offers more faithful token attribution by consistently emphasizing answer-bearing tokens, which helps explain its stronger empirical performance.

\section{Related Work}
\label{rel}

\paragraph{Retrieval Augmented Generation for Knowledge-Intensive QA}
Retrieval Augmented Generation (RAG) combines parametric LLM knowledge with an external corpus to improve factuality and coverage for knowledge-intensive tasks \cite{20nips_ragnlp, 20icml_realm, 24arxiv_nlprag}. A common design is to retrieve top-$k$ passages and condition generation on them, enabling explicit grounding and easy corpus updates compared to purely parametric memorization. In parallel, retrieval-enhanced language modeling approaches such as kNN-LM \cite{20iclr_knnlm} and RETRO~\cite{22icml-retro} also demonstrate that external memories can improve prediction by injecting neighbors from large datastores. Our work aligns with this direction of grounding generation in explicit evidence, but focuses on a different bottleneck: \emph{how to integrate retrieved information into the model efficiently and robustly}.

\paragraph{Retrieval Fusion}
The predominant way to use retrieval in modern LLM-based QA is \emph{in-context injection}, where retrieved passages are appended to the prompt \cite{20nips_ragnlp,24naacl_replug}. While conceptually simple and compatible with black-box LLMs, this approach grows context length with the amount of evidence, leading to increased inference cost due to quadratic self-attention complexity in Transformers \cite{vaswani2017attention}. 
An alternative to long-context prompting is to internalize external knowledge into model parameters, through continued pretraining, fine-tuning, or modular add-ons. Adapter-based knowledge infusion methods such as K-Adapter~\cite{21acl_kadpater} and parameter-efficient fine-tuning techniques such as LoRA~\cite{22iclr_lora, prag, dyprag} can inject domain knowledge while keeping most backbone weights frozen. However, moving knowledge into parameters complicates rapid updates and weakens explicit evidence traceability. Our method preserves a non-parametric, updatable corpus and inspectable evidence selection while learning a lightweight integration interface that avoids full long-context processing.

\section{Conclusion}
\label{con}

We introduced a new latent-based fusion method, called ReFilter, for retrieval-augmented generation that employs token-level evidence filtering and fusion.
This approach bypasses lengthy context windows, mitigates noise sensitivity, and guides answer generation without overloading the backbone model. 
Experiments show that our ReFilter maintains efficiency, robustness to noisy retrieval, and traceable corpus use across both in-domain and out-of-domain settings.

\bibliographystyle{ACM-Reference-Format}
\bibliography{sample-sigconf}

\newpage
\appendix
\section{Appendix}
\label{app}

\subsection{Notations}
\label{notations}
The notations used in this paper are summarized in Table~\ref{tab:notations_part1} and Table~\ref{tab:notations_part2}.

\begin{table}[htbp]
\centering
\caption{Notations used in the ReFilter method (Part 1).}
\label{tab:notations_part1}
\renewcommand{\arraystretch}{1.0} 
\begin{tabularx}{1.0\linewidth}{@{} l X @{}}
\toprule
\textbf{Notation} & \textbf{Definition} \\
\midrule
\multicolumn{2}{l}{\textit{\textbf{Dimensions and Constants}}} \\
\midrule
$B$ & Batch size \\
$P$ & Input sequence length \\
$L$ & Total number of layers in the LLM \\
$s$ & Fixed token length per chunk (denoted as \texttt{chunk\_len}) \\
$k$ & Number of retrieval chunks for each query \\
$N$ & Total tokens in the retrieval pool per query ($N = k \times s$) \\
$T$ & Length of target answer sequence \\
\midrule
\multicolumn{2}{l}{\textit{\textbf{Query and Retrieval}}} \\
\midrule
$Q$ & A batch of queries $\{q_1, \dots, q_B\}$ \\
$q_b$ & The $b$-th query in the batch, where $b \in \{1, \dots, B\}$ \\
$\mathcal{C}$ & External knowledge corpus \\
$\mathcal{C}_k(q_b)$ & Set of retrieval chunks for query $q_b$: 
$\{c_{b,1},\dots,c_{b,k}\}$ \\
$c_{b,i}$ & The $i$-th retrieval chunk for query $q_b$, where $i \in \{1, \dots, k\}$ \\
$\texttt{chunk\_len}$ & Fixed token length per chunk (also denoted as $s$) \\
\midrule

\multicolumn{2}{l}{\textit{\textbf{Context Encoder}}} \\
\midrule
$E(\cdot)$ & External Transformer encoder \\
$f_{b,i}$ & Text feature of chunk $c_{b,i}$ \\
$d_e$ & Feature dimension of the external encoder \\
$d_m$ & Projected feature dimension \\
$W_p$ & Projection matrix mapping $d_e \to D$ \\
$C_{b,i}$ & Projected context feature of chunk $c_{b,i}$ \\
$C$ &  Context embeddings (projected token pool): $C \in \mathbb{R}^{B \times N \times d_m}$ \\
\midrule

\multicolumn{2}{l}{\textit{\textbf{Gated Filter}}} \\
\midrule
$\mathbf{H}$ & LLM hidden states: $\mathbf{H} \in \mathbb{R}^{B \times P \times L \times D}$ \\
$p_b$ & Injection position index for the $b$-th query \\
$\mathbf{p}$ & A batch of injection positions: $\mathbf{p} \in \mathbb{N}^B$ \\
$h_b$ & Decision State: LLM hidden state of query $q_b$ at injection position ($p_b,L$) \\
$C_{b,j}$ & Token embedding vector for batch $b$, token position $j$ (element of $C$) \\
$g_{b,j}$ & Concatenated feature for gating: $g_{b,j} = [ C_{b,j} ; h_b ] \in \mathbb{R}^{2D}$ \\
$W_g, a_g$ & Learnable weights and bias of the dynamic gate scorer \\
$\gamma_{b,j}$ & Dynamic Gate value: $\sigma(W_g^\top g_{b,j} + a_g)$ \\
$\mu_j$ & Learnable position mask: $\mu_j \in \mathbb{R}$, $j = 1, \dots, N$ \\
$\mu$ & Set of position masks: $\mu = \{\mu_1, \ldots, \mu_N\}$ \\
$W_t$ & Final token weights: derived from $\mu_j \cdot \gamma_{b,j}$ \\
$\hat{C}$ & Weighted token features: $\hat{C} = W_t \cdot C \in \mathbb{R}^{B \times N \times d_m}$ \\
$\sigma(\cdot)$ & Sigmoid activation function \\
\bottomrule
\end{tabularx}
\end{table}

\begin{table}[t]
\centering
\caption{Notations used in the ReFilter method (Part 2).}
\label{tab:notations_part2}
\begin{tabularx}{\linewidth}{@{} l X @{}}
\toprule
\textbf{Notation} & \textbf{Definition} \\
\midrule
\multicolumn{2}{l}{\textit{\textbf{Token Fusion}}} \\
\midrule
$\hat{C}_{b,j}$ & Weighted token feature for batch $b$, token position $j$ (element of $\hat{C}$) \\
$r_b$ & Aggregated evidence representation: $r_b = \mathrm{LayerNorm}\left(\sum_{j=1}^{N} \mathrm{Dropout}(\hat{C}_{b,j})\right) \in \mathbb{R}^D$ \\
$\alpha$ & Learnable scaling factor for injection strength \\
$\mathbf{H}_{b,p_b,L,:}$ & Hidden state of query $q_b$ at injection position $p_b$ and fusion layer $L$: $\mathbf{H}_{b,p_b,L,:} \in \mathbb{R}^D$ \\
$\hat{\mathbf{H}}_{b,p_b,L,:}$ & Updated hidden state after injection: $\hat{\mathbf{H}}_{b,p_b,L,:} = \mathbf{H}_{b,p_b,L,:} + \alpha \cdot r_b \in \mathbb{R}^D$ \\
$\mathrm{Dropout}(\cdot)$ & Dropout operation \\
$\mathrm{LayerNorm}(\cdot)$ & Layer normalization \\
\midrule
\multicolumn{2}{l}{\textit{\textbf{Training Objective}}} \\
\midrule
$y$ & Target answer tokens $y = [y_1, \dots, y_T]$ \\
$y_t$ & Target answer token at position $t$ \\
$\mathcal{L}_{\text{NLL}}$ & Negative log-likelihood loss \\
$\mathcal{L}_{\text{gate}}$ & Gate sparsity regularization \\
$\lambda$ & Balancing factor between $\mathcal{L}_{\text{NLL}}$ and $\mathcal{L}_{\text{gate}}$ \\
$\mathcal{L}$ & Total loss: $\mathcal{L} = \mathcal{L}_{\text{NLL}} + \lambda \mathcal{L}_{\text{gate}}$ \\
\bottomrule
\end{tabularx}
\end{table}

\subsection{Scaling to Large Models}
To evaluate the scalability of our approach, we further conduct experiments on larger backbone models with 32B and 72B parameters and trained on the training split of the target dataset. As shown in the Table~\ref{tab:more_power}, our method consistently outperforms the corresponding baseline models across both scales. This demonstrates that the performance gains of our approach are preserved as model size increases, indicating good scalability and robustness to backbone scaling.

\begin{table*}[t]
\centering
\caption{Results on more powerful LLM backbones of ReFilter.}
\label{tab:more_power}
\begin{tabular}{lcccccc}
\toprule
\textbf{Backbone} & \textbf{Method} & \textsc{2WQA} & \textsc{HPQA} & \textsc{PopQA} & \textsc{CWQ} & \textbf{AVG.} \\
\midrule
\multirow{3}{*}{Qwen2.5-32B-Instruct} & Vanilla & 33.11	&31.46	&25.51	&50.29	& 35.09  \\
& S-RAG & 33.87	&32.98	&27.46	&51.20	&36.38  \\
& Ours & 35.03	&35.26	&27.16	&51.86	&\textbf{37.33}\\
\midrule
\multirow{3}{*}{Qwen2.5-72B-Instruct} & Vanilla & 34.03	&35.54	&27.14	&53.05	&37.44 \\
& S-RAG & 31.88	&35.71	&35.11	&52.57	&38.82 \\
& Ours& 36.98	&35.86	&35.03	&54.42	&\textbf{40.57}\\

\bottomrule
\end{tabular}
\end{table*}

\subsection{Effect of Fusion Depth}
We study where to fuse retrieved information into the backbone model by comparing fusion at the last layer, the last three layers, and the last five layers. As shown in the Table~\ref{tab:fusion_depth}, fusing retrieval information only at the final layer consistently yields the best performance across datasets. Moreover, this design is more efficient, as it avoids repeated fusion operations across multiple layers. These findings suggest that late-stage fusion is sufficient to effectively inject useful retrieval signals, while deeper fusion introduces redundant computation without clear performance benefits.

\begin{table*}[t]
\centering
\caption{Results on different layer fusion.}
\label{tab:fusion_depth}
\begin{tabular}{lcccccc}
\toprule
\textbf{Backbone} & \textbf{Method} & \textsc{2WQA} & \textsc{HPQA} & \textsc{PopQA} & \textsc{CWQ} & \textbf{AVG.} \\
\midrule
\multirow{3}{*}{LLaMA-3.2-1B-Instruct} & last layer & 26.87	&25.28	&23.38	&35.98	&\textbf{27.88}  \\
& last three layers & 27.12	&24.47	&21.26	&35.01	&26.97  \\
& last five layers & 26.12	&21.77	&23.48	&34.25	&26.41 \\
\midrule
\multirow{3}{*}{LLaMA-3-8B-Instruct} & last layer &38.42	&35.32	&27.38	&46.49	&\textbf{36.90}  \\
& last three layers & 39.26	&38.13	&26.30	&41.92	&36.40 \\
& last five layers& 39.10	&29.34	&25.29	&40.88	&33.65\\
\midrule
\multirow{3}{*}{Qwen2.5-1.5B-Instruct} & last layer & 27.43	&19.39	&19.02	&35.02	&\textbf{25.22} \\
& last three layers & 29.47	&20.37	&18.38	&32.66	&25.22 \\
& last five layers& 27.01	&16.46	&16.29	&38.49	&24.56\\
\midrule
\multirow{3}{*}{Qwen2.5-14B-Instruct} & last layer & 37.75	&27.40	&29.31	&49.89	&36.09 \\
& last three layers & 39.23	&28.46	&30.37	&50.32	&\textbf{37.10} \\
& last five layers& 40.36	&29.67	&27.31	&49.21	&36.64\\
\bottomrule
\end{tabular}
\end{table*}

\subsection{Prompt templates}
In this section, we present the prompt templates used in our experiments. During inference, we employ the prompt in Table~\ref{tab:prompt_general} for general-domain QA datasets and the prompt in Table~\ref{tab:prompt_medical} for biomedical QA datasets.

\begin{table*}[h]
\centering
\begin{tabular}{@{} p{0.65\linewidth} @{}}
\toprule
\textbf{System Prompt:} You are a helpful and precise assistant. Answer the user's question concisely using your knowledge. \\
\textbf{User Prompt:} Output only a single entity or a short phrase, without explanations. \\
Question: \{Question\} \\
Answer: \\
\bottomrule
\end{tabular}
\caption{Prompt template used for general-domain QA benchmarks.}
\label{tab:prompt_general}
\end{table*}

\begin{table*}[h]
\centering
\begin{tabular}{@{} p{0.65\linewidth} @{}}
\toprule
\textbf{System Prompt:} 
You are a medical expert. Answer the medical questions based on the given options. \\
Output only the option letter (A, B, C, or D), without any additional text. \\
\textbf{User Prompt:} 
Given the question, choose one of the options as your answer. Output only A, B, C, or D. \\
Question: \{Question\} \\
Options: \\
A. \{Option A\} \\
B. \{Option B\} \\
C. \{Option C\} \\
D. \{Option D\} \\
Answer: \\
\bottomrule
\end{tabular}
\caption{Prompt template used for medical QA benchmarks.}
\label{tab:prompt_medical}
\end{table*}
\end{document}